\documentclass{article}
\usepackage{float}
\usepackage{tabularx}
\usepackage{PRIMEarxiv}
\usepackage{natbib}
\usepackage[utf8]{inputenc} 
\usepackage[T1]{fontenc}    
\usepackage{hyperref}       
\usepackage{url}            
\usepackage{booktabs}       
\usepackage{amsfonts}       
\usepackage{nicefrac}       
\usepackage{microtype}      
\usepackage{lipsum}
\usepackage{fancyhdr}       
\usepackage{graphicx}       
\graphicspath{{media/}}     
\usepackage{longtable}
\usepackage{float}
\usepackage{enumitem}
\usepackage{amsmath}
\usepackage{graphicx}
\usepackage{rotating}
\usepackage{array}
\usepackage{makecell}
\usepackage{algorithm2e}

\title{The Million-Label NER: Breaking Scale Barriers with GLiNER bi-encoder
}
\author{
\textbf{Ihor Stepanov$^{1}$, Mykhailo Shtopko$^{1}$, Dmytro Vodianytskyi$^{1}$,} Oleksandr Lukashov$^{1}$\\
$^{1}$Knowledgator Engineering, Kyiv, Ukraine \\
\textbf{Correspondence:} \texttt{ingvarstep@knowledgator.com}, \texttt{mykhailoshtopko@knowledgator.com}
}

\begin{document}
\maketitle
\begin{abstract}
This paper introduces GLiNER-bi-Encoder, a novel architecture for Named Entity Recognition (NER) that harmonizes zero-shot flexibility with industrial-scale efficiency. While the original GLiNER framework offers strong generalization, its joint-encoding approach suffers from quadratic complexity as the number of entity labels increases. Our proposed bi-encoder design decouples the process into a dedicated label encoder and a context encoder, effectively removing the context-window bottleneck. This architecture enables the simultaneous recognition of thousands—and potentially millions—of entity types with minimal overhead. Experimental results demonstrate state-of-the-art zero-shot performance, achieving 61.5\% Micro-F1 on the CrossNER benchmark. Crucially, by leveraging pre-computed label embeddings, GLiNER-bi-Encoder achieves up to a 130$\times$ throughput improvement at 1024 labels compared to its uni-encoder predecessors. Furthermore, we introduce GLiNKER, a modular framework that leverages this architecture for high-performance entity linking across massive knowledge bases such as Wikidata.
\end{abstract}


\section{Introduction}\label{sec:introduction}

Information extraction (IE) remains a fundamental challenge in natural language processing, with critical applications spanning scientific research \citep{Hong2021ChallengesAA}, healthcare \citep{Yazdani2024GLiNERBiomed}, finance \citep{Skalicky2022BusinessDI}, and public administration \citep{Siciliani2023OIE4PAOI}. The exponential growth of unstructured text data across these domains demands IE systems that can efficiently process large volumes while maintaining high accuracy—particularly in high-stakes applications where extraction errors can have serious consequences. Moreover, the dynamic nature of real-world applications requires models that can rapidly adapt to new entity types and domains without extensive retraining.

Named Entity Recognition (NER), as a cornerstone IE task, has evolved significantly from early rule-based systems to modern neural approaches. While rule-based methods \citep{petasis2001using} required extensive manual effort and lacked generalization capabilities, statistical approaches like Hidden Markov Models \citep{seymore1999learning} and Conditional Random Fields \citep{mcdonald2005identifying} introduced data-driven learning but struggled with complex contextual dependencies. The transformer revolution, initiated by BERT \citep{devlin2018bert}, transformed NER by enabling bidirectional context understanding, leading to substantial performance improvements across benchmarks.

Recent advances have explored two primary paradigms for zero-shot and few-shot NER. Generative approaches, exemplified by large language models like GPT-4 \citep{openai2024gpt4}, frame NER as a text generation task, offering flexibility in handling diverse entity types through natural language prompts. However, these methods suffer from computational inefficiency—generating tokens for entities already present in the text—and often struggle with structured output consistency, leading to hallucinations and format violations that compromise reliability in production systems.

In contrast, discriminative approaches have shown promise in maintaining efficiency while achieving competitive performance. GLiNER \citep{zaratiana2023gliner} introduced a unified token classification architecture that encodes both text and entity type descriptions via a single transformer, enabling zero-shot entity recognition through attention-based label-text interactions. While effective, this approach faces scalability challenges when handling large label sets, as all labels must be processed jointly with each input text, resulting in quadratic complexity with respect to the number of entity types.

In this work, we present a bi-encoder architecture that addresses these limitations by decoupling text and entity type encodings. Our approach separates the encoding process into two specialized components: a text encoder that processes input sequences to generate span representations, and a label encoder that independently embeds entity type descriptions. This separation offers several key advantages: (1) entity type embeddings can be pre-computed and cached, significantly reducing inference time for large label sets; (2) leveraging pre-trained sentence transformers as label encoders provides superior semantic understanding for unseen entity types; and crucially, (3) the decoupled design enables the model to efficiently handle thousands or even millions of entity types simultaneously, as label embeddings are computed independently and can be stored in efficient vector databases for similarity search.

This scalability is particularly important for real-world applications such as biomedical entity linking, where ontologies like UMLS \citep{bodenreider2004unified} contain over 4 million concepts, or enterprise knowledge extraction systems that must recognize entities across numerous taxonomies. Unlike joint encoding approaches that face memory and computational constraints when scaling beyond hundreds of labels, our bi-encoder architecture maintains constant memory usage for text encoding regardless of the number of entity types, with label retrieval complexity reduced to an efficient nearest-neighbor search.

Experimental results demonstrate that our bi-encoder approach achieves state-of-the-art performance on zero-shot NER benchmarks while offering 2-3× faster inference than uni-encoder GLiNER with dozens of entity types, with the performance gap widening to 10-100× when scaling to thousands of entity types. Moreover, the model exhibits strong generalization to specialized domains without fine-tuning, making it particularly suitable for real-world deployment scenarios where entity types evolve dynamically, and scale requirements are demanding.

Beyond standalone named entity recognition, the bi-encoder architecture's ability to efficiently handle large entity type sets makes it particularly well-suited for entity linking tasks. Entity linking extends NER by disambiguating mentions to specific knowledge base entities, requiring candidate retrieval and entity disambiguation—operations that benefit directly from precomputed embeddings and efficient similarity search. To facilitate practical deployment of GLiNER-bi-Encoder in entity linking scenarios, we have developed GLiNKER, a modular DAG-based framework that integrates bi-encoder models into multi-stage linking pipelines combining mention extraction, candidate generation from knowledge bases, and linking entities.

\section{Methods}\label{sec:methods}

Zero-shot NER requires computing compatibility between text spans and arbitrary entity type definitions. The original GLiNER architecture addresses this by jointly encoding entity types with the input text and processing them with a single transformer. While this enables rich cross-attention between labels and context, it couples computational cost to label count—processing $n$ tokens against $m$ entity types requires self-attention over sequences of length $\mathcal{O}(n + m)$, with complexity $\mathcal{O}((n + m)^2)$.

The bi-encoder architecture restructures this computation by observing that entity type semantics are largely independent of input text. A description like ``person'' or ``chemical compound'' can be encoded once and reused across all inputs. This enables architectural separation: independent encoders process text and labels in parallel, combining embeddings only at the final scoring stage. Text encoding becomes $\mathcal{O}(n^2)$ independent of label count, while label encoding can be performed offline and cached. With pre-computed embeddings, inference complexity is effectively $\mathcal{O}(n^2)$ regardless of whether the system recognizes 10 or 10,000 entity types.

This separation also enables heterogeneous encoder selection—pairing text encoders optimized for contextual understanding with label encoders optimized for semantic similarity. To address the potential loss of cross-modal interaction, an optional cross-attention fusion layer can exchange information between representations after initial encoding, offering a tunable trade-off between accuracy and efficiency.

\subsection{Architecture Overview}

The bi-encoder architecture employs two specialized, independent transformers. This separation is the fundamental driver of the model's scalability, as it allows for heterogeneous backbone selection (e.g., pairing a fast text encoder with a semantically rich sentence-transformer for labels).

\begin{figure}[H]
    \centering
    \includegraphics[width=1\linewidth]{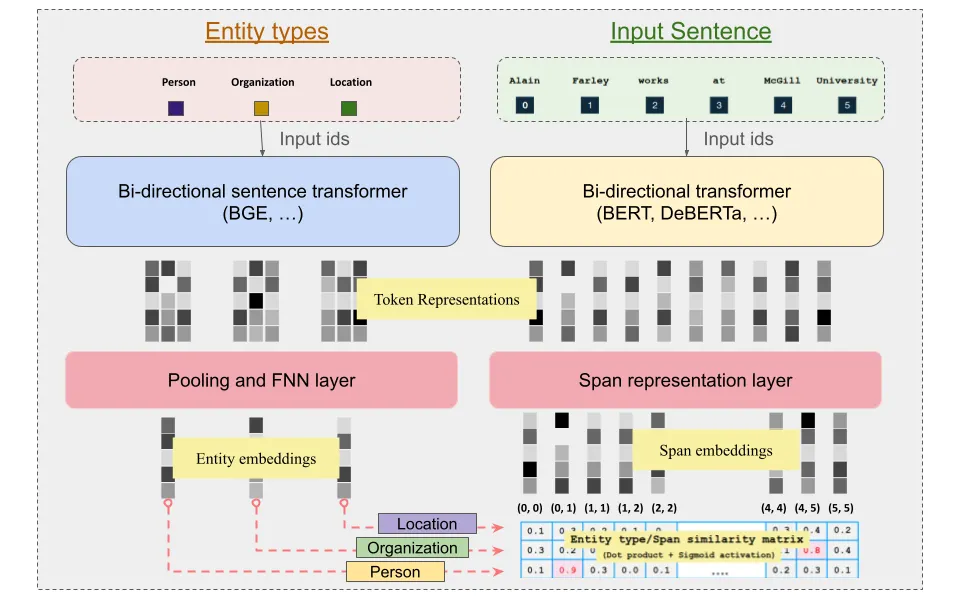}
    \caption{GLiNER architecture variants. Left: Uni-encoder processes entity types and text jointly. Right: Bi-encoder uses separate encoders with optional cross-attention fusion.}
    \label{fig:architecture}
\end{figure}

\subsection{Uni-Encoder Architecture}

Before jumping into the details of the bi-encoder architecture, let's discuss the essence of the original GLiNER architecture. The uni-encoder processes entity types and input text within a single transformer. Entity types are prepended to the input using special tokens:
\begin{equation}
    \mathbf{X} = [\texttt{[ENT]}~t_1~\texttt{[ENT]}~t_2~\cdots~\texttt{[SEP]}~w_1~w_2~\cdots~w_L]
\end{equation}
where $t_i$ denotes entity type tokens and $w_j$ denotes input words. The transformer produces contextualized embeddings:
\begin{equation}
    \mathbf{H} = \text{Transformer}(\mathbf{X}) \in \mathbb{R}^{N \times D}
\end{equation}

Entity type embeddings $\mathbf{E} \in \mathbb{R}^{C \times D}$ are extracted at \texttt{[ENT]} token positions, and word embeddings $\mathbf{W} \in \mathbb{R}^{L \times D}$ are extracted using a word-to-subtoken mapping that selects the first subtoken of each word.

\subsection{Bi-Encoder Architecture}

The bi-encoder employs separate transformers for entity types and input text:
\begin{align}
    \mathbf{E} &= \text{LabelEncoder}(\mathbf{T}) \in \mathbb{R}^{C \times D} \\
    \mathbf{H} &= \text{TextEncoder}(\mathbf{X}) \in \mathbb{R}^{N \times D}
\end{align}
where $\mathbf{T}$ contains tokenized entity type descriptions. Word embeddings $\mathbf{W}$ are extracted from $\mathbf{H}$ as in the uni-encoder case. The label embeddings are expanded across the batch: $\mathbf{E}_{\text{batch}} = \text{expand}(\mathbf{E}, B) \in \mathbb{R}^{B \times C \times D}$.

This design enables caching of entity type representations, significantly improving inference efficiency when the same entity types are queried repeatedly.

\subsubsection{Cross-Attention Fusion}

Both architectures optionally incorporate bidirectional cross-attention to exchange information between text and entity representations:
\begin{equation}
    \mathbf{W}', \mathbf{E}' = \text{CrossFuser}(\mathbf{W}, \mathbf{E}, \mathbf{M}_{\text{text}}, \mathbf{M}_{\text{labels}})
\end{equation}
The CrossFuser applies multi-head attention in both directions according to a configurable schema, allowing entity-aware text representations and text-aware entity representations. This fusion occurs after initial encoding and before span/token scoring.

\subsection{Span-Level Prediction}

Span-level models enumerate all candidate spans up to a maximum width $K$ and score each against entity types.

\subsubsection{Span Representation}

Word embeddings are projected through separate MLPs for span boundaries:
\begin{align}
    \mathbf{H}_{\text{start}} &= \text{MLP}_{\text{start}}(\mathbf{W}) \in \mathbb{R}^{B \times L \times D} \\
    \mathbf{H}_{\text{end}} &= \text{MLP}_{\text{end}}(\mathbf{W}) \in \mathbb{R}^{B \times L \times D}
\end{align}

For each span $(i, i+k)$ where $k \in [0, K-1]$:
\begin{equation}
    \mathbf{s}_{i,k} = \text{MLP}_{\text{out}}\left(\text{ReLU}\left([\mathbf{h}_{\text{start}}^{i} \| \mathbf{h}_{\text{end}}^{i+k}]\right)\right)
\end{equation}
yielding span representations $\mathbf{S} \in \mathbb{R}^{B \times L \times K \times D}$.

\subsubsection{Scoring}

Entity type embeddings are projected and scored against spans via:
\begin{equation}
    \mathbf{O} = \mathbf{S} \cdot \text{MLP}_{\text{prompt}}(\mathbf{E})^\top \in \mathbb{R}^{B \times L \times K \times C}
\end{equation}
where $O_{b,i,k,c}$ represents the score for span $(i, i+k)$ belonging to entity class $c$ in batch example $b$.

\subsection{Token-Level Prediction}

Token-level models classify each word position independently, predicting BIO-style tags for each entity type.

\subsubsection{Token Scoring}

A scoring layer computes compatibility between word embeddings and entity types:
\begin{equation}
    \mathbf{O} = \text{Scorer}(\mathbf{W}, \mathbf{E}) \in \mathbb{R}^{B \times L \times C \times 3}
\end{equation}
The final dimension encodes \texttt{[start, end, inside]} probabilities for BIO tagging.

\subsubsection{Span Extraction from Tokens}

Entity spans are extracted by identifying consistent BIO sequences:
\begin{enumerate}
    \item Identify positions where $P(\text{start}) > \tau$ and $P(\text{end}) > \tau$
    \item For each (start, end) pair of the same class where $\text{start} \leq \text{end}$:
    \item Validate that all intermediate positions satisfy $P(\text{inside}) > \tau$
    \item Accept spans where inside predictions span the full interval
\end{enumerate}

\subsubsection{Optional Span Representation}

Token-level models can optionally compute explicit span representations for extracted spans:
\begin{equation}
    \mathbf{s}_{(i,j)} = \text{SpanRep}(\mathbf{W}, i, j)
\end{equation}
These representations enable a secondary span-level classification loss:
\begin{equation}
    \mathcal{L}_{\text{total}} = \lambda_{\text{token}} \mathcal{L}_{\text{token}} + \lambda_{\text{span}} \mathcal{L}_{\text{span}}
\end{equation}

\subsection{Training}

\subsubsection{Loss Function}

We use focal loss \cite{lin2018focallossdenseobject} to handle class imbalance:
\begin{equation}
    \mathcal{L}_{\text{focal}} = -\alpha(1-p_t)^{\gamma}\log(p_t)
\end{equation}
where $p_t = \sigma(o)$ for positive examples and $p_t = 1 - \sigma(o)$ for negatives.

\subsubsection{Negative Sampling}

Given the quadratic number of candidate spans, we apply stochastic negative masking:
\begin{equation}
    m_{i,k,c} = \begin{cases}
        1 & \text{if } y_{i,k,c} = 1 \\
        \text{Bernoulli}(\rho) & \text{if } y_{i,k,c} = 0
    \end{cases}
\end{equation}
where $\rho$ controls the proportion of negative examples retained. Masking strategies include global (uniform), label-wise, and span-wise sampling.

\subsection{Inference}

\subsubsection{Greedy Span Decoding}

Given logits $\mathbf{O}$, we decode entities through threshold filtering and greedy selection:

\begin{enumerate}
    \item \textbf{Filter}: Extract candidates $\mathcal{C} = \{(s, e, c, p) : \sigma(O_{s,k,c}) > \tau\}$ where $e = s + k$
    \item \textbf{Sort}: Order candidates by probability descending
    \item \textbf{Select}: Greedily accept spans that don't conflict with already-selected spans
    \item \textbf{Output}: Return spans sorted by position
\end{enumerate}

\subsubsection{Overlap Strategies}

The conflict detection function supports multiple NER paradigms:

\textbf{Flat NER}: Two spans conflict if they overlap:
\begin{equation}
    \text{conflict}(s_1, e_1, s_2, e_2) = \max(s_1, s_2) \leq \min(e_1, e_2)
\end{equation}

\textbf{Multi-label NER}: Spans of different types may overlap:
\begin{equation}
    \text{conflict}((s_1, e_1, c_1), (s_2, e_2, c_2)) = (c_1 = c_2) \land \text{overlap}(s_1, e_1, s_2, e_2)
\end{equation}

\textbf{Nested NER}: Proper containment is permitted:
\begin{equation}
    \text{conflict} = \text{overlap} \land \neg\text{contained}
\end{equation}
where $\text{contained}((s_1, e_1), (s_2, e_2)) = (s_1 \leq s_2 \land e_2 \leq e_1) \lor (s_2 \leq s_1 \land e_1 \leq e_2)$.

\subsection{Data}\label{sec:data}

\paragraph{Pre-training corpus:}
The pre-training dataset consists of 8M samples from \texttt{m-a-p/FineFineWeb} for Large, Base, and Small variants. The Edge variant was trained on an extended corpus of 10M samples. Each text was annotated with GPT-4o to generate true candidate labels for entity recognition training.

\paragraph{Post-training corpus:}
The post-training dataset consists of 40k high-quality samples with sequences up to 2048 tokens, designed to refine model performance on longer contexts and improve generalization across diverse entity types.

\subsection{Model Training}

\subsubsection{Model Variants}
We trained four model variants with different capacity-efficiency trade-offs based on the ModernBERT encoder architecture:

\begin{itemize}
    \item \textbf{Large:} 400M parameters (ettin-encoder-400m), 28 layers, 16 attention heads, hidden size 1024, paired with BAAI/bge-base-en-v1.5 label encoder (768-dim)
    \item \textbf{Base:} 150M parameters (ettin-encoder-150m), 22 layers, 12 attention heads, hidden size 768, paired with BAAI/bge-small-en-v1.5 label encoder (384-dim)
    \item \textbf{Small:} 68M parameters (ettin-encoder-68m), 19 layers, 8 attention heads, hidden size 512, paired with sentence-transformers/all-MiniLM-L12-v2 label encoder (384-dim)
    \item \textbf{Edge:} 32M parameters (ettin-encoder-32m), 10 layers, 6 attention heads, hidden size 384, paired with sentence-transformers/all-MiniLM-L6-v2 label encoder (384-dim)
\end{itemize}

All variants share the ModernBERT architecture configuration with position-free embeddings (sans\_pos), local attention window of 128 tokens, global attention every 3 layers, RoPE theta of 160,000, maximum position embeddings of 7,999, and vocabulary size of 50,368 tokens.

\subsubsection{Training Stages}

\paragraph{Pre-Training:}
Initial training on the pre-training corpus for one complete epoch with the following configuration:
\begin{itemize}
    \item Dataset size: 8M samples (Large/Base/Small), 10M samples (Edge)
    \item Maximum sequence length: 1024 tokens
    \item Focal loss $\alpha$ parameter: 0.7
    \item Focal loss $\gamma$ parameter: 2.0
    \item Training steps: 500k (Large), 250k (Base/Small), 312.5k (Edge)
    \item Batch size: 16 (Large), 32 (Base/Small/Edge)
\end{itemize}

\paragraph{Post-Training:}
Final stage training on the post-training corpus for one complete epoch with increased context length and adjusted loss parameters:
\begin{itemize}
    \item Maximum sequence length: 2048 tokens
    \item Focal loss $\alpha$ parameter: 0.8
    \item Focal loss $\gamma$ parameter: 2.0
    \item Loss reduction: sum aggregation
\end{itemize}

\subsubsection{Optimization}
We employed the AdamW optimizer with differential learning rates for the encoder and other components:
\begin{itemize}
    \item Encoder learning rate: $1 \times 10^{-5}$
    \item Other components learning rate: $3 \times 10^{-5}$
    \item Weight decay (encoder): 0.01
    \item Weight decay (other): 0.01
    \item Gradient clipping: maximum norm of 10.0
    \item Learning rate scheduler: cosine annealing
    \item Warmup ratio: 0.1 (10\% of total training steps)
\end{itemize}

\subsubsection{Architectural Configuration}
The models were configured with the following entity recognition parameters:
\begin{itemize}
    \item Span detection mode: MarkerV0
    \item Subtoken pooling strategy: first token
    \item Maximum entity types per batch: 100
    \item Maximum span width: 12 tokens
    \item Dropout rate: 0.35
    \item RNN component: enabled
\end{itemize}

\subsubsection{Training Details}
Models were trained with the following data processing and regularization settings:
\begin{itemize}
    \item Word splitter: whitespace tokenization
    \item Entity type shuffling: enabled during training
    \item Maximum negative type ratio: 1.0
    \item Label smoothing: 0.0 (disabled)
\end{itemize}

All models utilized pre-trained label encoders for encoding entity type, which were fine-tuned jointly with the main encoder during training. The focal loss formulation helps address class imbalance by focusing on hard examples while down-weighting easy negatives, particularly important for NER, where non-entity tokens vastly outnumber entity tokens.
\section{Evaluation}\label{seq:results}

The model was compared with other GLiNER-type models. All models were evaluated on cross-domain NER benchmarks in a zero-shot setting. We documented Micro-F1 scores across all datasets. The datasets encompass diverse domains including AI, Literature, Politics, Science, Movies, and other (see Tables~\ref{tab:gliner-bi-res} and~\ref{tab:gliner-uni-res}). A threshold of 0.4 was applied to filter predicted entities, and all evaluations were conducted at the span level.

Inference speed is measured on a single NVIDIA H100 GPU with batch size 1. We test across label counts $L \in \{1, 2, 4, 8, 16, 32, 64, 128, 256, 512, 1024\}$ and input lengths $T \in \{64, 256, 512\}$ tokens. For each $(L,T)$ configuration, we execute 10 forward passes and report average throughput in examples per second.

\subsection{Model Configuration and Performance}

We evaluated four bi-encoder model variants ranging from 60M to 530M parameters, utilizing text encoders from the Ettin suite \citep{weller2025seqvsseqopen} paired with specialized label encoders from sentence-transformers and BGE families \citep{bge_embedding}. Table~\ref{tab:gliner-bi-overview} presents the complete model specifications and performance metrics.

\begin{table}[H]
\centering
\caption{GLiNER-bi-V2 Models Overview}
\label{tab:gliner-bi-overview}
\scriptsize
\begin{tabular}{lllllll}
\toprule
\textbf{Model name} & \textbf{Params} & \textbf{Encoder} & \textbf{Labels Encoder} & \textbf{Avg. CrossNER} & \textbf{Inference Speed} & \textbf{Inference Speed} \\
 &  &  &  & \textbf{Benchmark} & \textbf{(H100, examples/s)} & \textbf{(pre-computed labels)} \\
\midrule
gliner-bi-edge-v2.0 & 60 M & ettin-encoder-32m & all-MiniLM-L6-v2 & 54.0\% & 13.64 & 24.62 \\
gliner-bi-small-v2.0 & 108 M & ettin-encoder-68m & all-MiniLM-L12-v2 & 57.2\% & 7.99 & 15.22 \\
gliner-bi-base-v2.0 & 194 M & ettin-encoder-150m & bge-small-en-v1.5 & 60.3\% & 5.91 & 9.51 \\
gliner-bi-large-v2.0 & 530 M & ettin-encoder-400m & bge-base-en-v1.5 & 61.5\% & 2.68 & 3.60 \\
\bottomrule
\end{tabular}
\end{table}

The bi-encoder architecture demonstrates consistent performance improvements with model scaling. On the CrossNER benchmark, performance increases from 54.0\% (edge) to 61.5\% (large), representing a 13.9\% relative improvement. Label pre-computation yields 1.34-1.91× speedup across all model sizes, with the most substantial gains observed in smaller models. The base variant (194M parameters) achieves 98\% of the large model's performance while operating 2.6× faster, establishing it as the optimal configuration for most deployment scenarios.

\subsection{Zero-shot NER Benchmarks}

Table~\ref{tab:gliner-bi-res} presents comprehensive evaluation results across 19 diverse NER datasets spanning biomedical, social media, news, and technical domains. Performance patterns reveal strong domain generalization capabilities, with WikiNeural (76.6-80.0\%), bc5cdr (68.5-73.0\%), and Broad Tweet Corpus (70.0-72.1\%) showing the highest F1 scores across model variants.

\begin{table}[H]
\centering
\caption{Comparison of \texttt{gliner-bi-v2.0} models on zero-shot NER benchmarks}
\label{tab:gliner-bi-res}
\small
\begin{tabular}{lcccc}
\toprule
\textbf{Dataset} & \textbf{gliner-bi-edge-v2.0} & \textbf{gliner-bi-small-v2.0} & \textbf{gliner-bi-base-v2.0} & \textbf{gliner-bi-large-v2.0} \\
\midrule
ACE 2004            & 26.4\% & 27.5\% & 28.9\% & 31.9\% \\
ACE 2005            & 26.2\% & 28.1\% & 30.0\% & 31.4\% \\
AnatEM              & 39.1\% & 43.6\% & 35.4\% & 39.5\% \\
Broad Tweet Corpus  & 70.0\% & 71.7\% & 72.1\% & 70.9\% \\
CoNLL 2003          & 61.6\% & 64.2\% & 65.6\% & 66.5\% \\
FabNER              & 22.4\% & 23.2\% & 24.3\% & 22.7\% \\
FindVehicle         & 35.6\% & 40.3\% & 40.6\% & 39.1\% \\
GENIA\_NER          & 50.1\% & 53.8\% & 56.8\% & 60.1\% \\
HarveyNER           & 15.0\% & 10.6\% & 12.6\% & 14.7\% \\
MultiNERD           & 64.6\% & 66.0\% & 68.0\% & 64.0\% \\
Ontonotes           & 31.4\% & 31.9\% & 33.3\% & 32.5\% \\
PolyglotNER         & 45.1\% & 46.3\% & 46.6\% & 46.8\% \\
TweetNER7           & 36.9\% & 40.9\% & 40.4\% & 41.7\% \\
WikiANN en          & 52.3\% & 54.0\% & 54.9\% & 56.3\% \\
WikiNeural          & 78.0\% & 79.9\% & 80.0\% & 76.6\% \\
bc2gm               & 58.1\% & 59.9\% & 62.7\% & 61.4\% \\
bc4chemd            & 45.8\% & 49.1\% & 53.6\% & 50.5\% \\
bc5cdr              & 68.5\% & 71.5\% & 73.0\% & 71.7\% \\
ncbi                & 65.9\% & 65.4\% & 65.2\% & 65.9\% \\
\midrule
\textbf{Average}    & \textbf{47.0\%} & \textbf{48.8\%} & \textbf{49.7\%} & \textbf{49.7\%} \\
\bottomrule
\end{tabular}
\vspace{1em}
\begin{tabular}{lcccc}
\toprule
\textbf{Dataset} & \textbf{gliner-bi-edge-v2.0} & \textbf{gliner-bi-small-v2.0} & \textbf{gliner-bi-base-v2.0} & \textbf{gliner-bi-large-v2.0} \\
\midrule
CrossNER\_AI         & 53.8\% & 54.7\% & 58.3\% & 57.4\% \\
CrossNER\_literature & 56.2\% & 62.6\% & 65.2\% & 63.2\% \\
CrossNER\_music      & 68.2\% & 72.3\% & 73.4\% & 74.0\% \\
CrossNER\_politics   & 68.7\% & 70.0\% & 70.8\% & 73.0\% \\
CrossNER\_science    & 63.2\% & 66.1\% & 68.0\% & 67.6\% \\
mit-movie            & 30.5\% & 35.2\% & 46.2\% & 51.0\% \\
mit-restaurant       & 37.1\% & 39.5\% & 40.3\% & 44.3\% \\
\midrule
\textbf{Average}     & \textbf{54.0\%} & \textbf{57.2\%} & \textbf{60.3\%} & \textbf{61.5\%} \\
\bottomrule
\end{tabular}
\end{table}

Biomedical benchmarks exhibit strong scaling effects, with GENIA\_NER improving from 50.1\% to 60.1\% (20\% relative gain) and bc4chemd from 45.8\% to 53.6\% between edge and base variants. Performance on social media datasets remains stable across model sizes, suggesting that informal text benefits less from increased model capacity. Challenging domains such as HarveyNER (10.6-15.0\%) and FabNER (22.4-24.3\%) show limited improvements with scaling, indicating inherent task difficulty rather than model limitations.

\subsection{Comparison with Uni-encoder Architecture}

Comparative evaluation against uni-encoder GLiNER models (Table~\ref{tab:gliner-uni-res}) reveals comparable performance between architectures on standard benchmarks. The bi-encoder GLiNER-bi-large-v2.0 achieves 61.5\% on CrossNER versus 60.9\% for gliner\_large-v2.5, while maintaining substantially different computational characteristics. On the broader benchmark suite, bi-encoder models achieve 49.7\% average F1 compared to 46.4-47.0\% for uni-encoder variants.

\begin{table}[H]
\centering
\caption{Comparison of \texttt{gliner-community v2.5} models on zero-shot NER benchmarks}
\label{tab:gliner-uni-res}
\small
\begin{tabular}{lccc}
\toprule
\textbf{Dataset} & \textbf{gliner\_small-v2.5} & \textbf{gliner\_medium-v2.5} & \textbf{gliner\_large-v2.5} \\
\midrule
ACE 2004            & 28.2\% & 25.9\% & 26.8\% \\
ACE 2005            & 28.5\% & 25.8\% & 27.2\% \\
AnatEM              & 39.7\% & 43.1\% & 38.1\% \\
Broad Tweet Corpus  & 65.0\% & 66.5\% & 66.7\% \\
CoNLL 2003          & 65.4\% & 64.9\% & 64.2\% \\
FabNER              & 22.4\% & 23.5\% & 25.1\% \\
FindVehicle         & 39.0\% & 41.1\% & 47.0\% \\
GENIA\_NER          & 48.5\% & 55.1\% & 46.9\% \\
HarveyNER           & 17.4\% & 18.9\% & 19.1\% \\
MultiNERD           & 57.0\% & 59.6\% & 50.9\% \\
Ontonotes           & 27.3\% & 26.1\% & 22.8\% \\
PolyglotNER         & 42.0\% & 43.1\% & 42.4\% \\
TweetNER7           & 38.9\% & 38.6\% & 38.9\% \\
WikiANN en          & 57.3\% & 58.1\% & 59.0\% \\
WikiNeural          & 71.4\% & 72.6\% & 73.0\% \\
bc2gm               & 51.4\% & 55.6\% & 52.5\% \\
bc4chemd            & 38.7\% & 43.9\% & 46.8\% \\
bc5cdr              & 65.9\% & 65.9\% & 68.7\% \\
ncbi                & 64.2\% & 64.3\% & 65.3\% \\
\midrule
\textbf{Average}    & \textbf{45.7\%} & \textbf{47.0\%} & \textbf{46.4\%} \\
\bottomrule
\end{tabular}
\vspace{1em}
\begin{tabular}{lccc}
\toprule
\textbf{Dataset (zero-shot)} & \textbf{gliner\_small-v2.5} & \textbf{gliner\_medium-v2.5} & \textbf{gliner\_large-v2.5} \\
\midrule
CrossNER\_AI         & 54.2\% & 54.7\% & 54.9\% \\
CrossNER\_literature & 62.5\% & 66.1\% & 65.2\% \\
CrossNER\_music      & 68.9\% & 72.9\% & 71.6\% \\
CrossNER\_politics   & 64.1\% & 68.1\% & 72.4\% \\
CrossNER\_science    & 63.5\% & 66.3\% & 63.9\% \\
mit-movie            & 50.3\% & 41.4\% & 53.0\% \\
mit-restaurant       & 39.8\% & 38.5\% & 45.0\% \\
\midrule
\textbf{Average}     & \textbf{57.6\%} & \textbf{58.3\%} & \textbf{60.9\%} \\
\bottomrule
\end{tabular}
\end{table}

Domain-specific comparisons show bi-encoder advantages on GENIA\_NER (60.1\% vs 46.9\%), MultiNERD (68.0\% vs 50.9\%), and Ontonotes (33.3\% vs 22.8\%). Uni-encoder models perform better on CoNLL 2003 (65.4\% vs 66.5\%) and certain specialized datasets. These patterns suggest that bi-encoder architectures excel when semantic similarity between labels and text is crucial, while uni-encoder models maintain advantages for well-established benchmark formats.

\subsection{Inference Speed and Scalability}

Table~\ref{tab:gliner-bi-inference-speed} presents inference speed measurements across varying label counts. The bi-encoder architecture demonstrates superior scaling characteristics compared to uni-encoder approaches, with performance degradation patterns revealing fundamental architectural differences.

\begin{table}[h]
\centering
\caption{Inference Speed: Samples per Second by Number of Labels (batch\_size = 1, on H100 GPU)}
\scriptsize
\label{tab:gliner-bi-inference-speed}
\begin{tabular}{lcccccccccccc}
\toprule
\multicolumn{1}{p{3.5cm}}{Model Name} & 1 & 2 & 4 & 8 & 16 & 32 & 64 & 128 & 256 & 512 & 1024 & \textbf{Average} \\
\midrule
gliner-bi-edge-v2.0 & 17.0 & 27.0 & 5.05 & 22.4 & 17.5 & 13.9 & 15.2 & 12.5 & 10.8 & 5.43 & 3.23 & \textbf{13.64} \\
gliner-bi-edge-v2.0 (pre-computed) & 19.3 & 25.0 & 28.2 & 32.6 & 31.0 & 32.6 & 22.2 & 22.7 & 22.2 & 16.9 & 18.3 & \textbf{24.62} \\
gliner-bi-small-v2.0 & 12.5 & 12.8 & 5.98 & 11.6 & 10.6 & 9.43 & 6.94 & 7.35 & 5.74 & 3.33 & 1.60 & \textbf{7.99} \\
gliner-bi-small-v2.0 (pre-computed) & 14.7 & 15.9 & 14.3 & 15.3 & 15.4 & 15.4 & 15.6 & 15.3 & 15.5 & 15.7 & 14.3 & \textbf{15.22} \\
gliner-bi-base-v2.0 & 8.13 & 8.62 & 4.85 & 8.00 & 7.52 & 6.76 & 5.71 & 5.21 & 4.64 & 3.21 & 2.30 & \textbf{5.91} \\
gliner-bi-base-v2.0 (pre-computed) & 9.52 & 10.2 & 9.80 & 9.95 & 10.0 & 9.93 & 8.93 & 6.71 & 9.35 & 9.71 & 10.5 & \textbf{9.51} \\
gliner-bi-large-v2.0 & 3.52 & 2.53 & 3.87 & 3.50 & 3.66 & 3.19 & 1.90 & 2.46 & 2.39 & 1.62 & 0.87 & \textbf{2.68} \\
gliner-bi-large-v2.0 (pre-computed) & 4.37 & 4.07 & 4.53 & 4.54 & 4.47 & 3.46 & 3.85 & 3.04 & 2.82 & 1.84 & 2.64 & \textbf{3.60} \\
\midrule
gliner\_small-v2.5 & 10.7 & 14.6 & 14.1 & 13.2 & 11.9 & 10.3 & 7.91 & 4.26 & 1.29 & 0.43 & 0.14 & \textbf{8.08} \\
gliner\_medium-v2.5 & 7.81 & 8.51 & 8.39 & 7.58 & 7.12 & 5.62 & 4.18 & 2.19 & 0.68 & 0.23 & 0.07 & \textbf{4.76} \\
gliner\_large-v2.5 & 2.89 & 3.28 & 3.29 & 2.90 & 2.61 & 2.33 & 1.71 & 1.12 & 0.31 & 0.09 & 0.03 & \textbf{1.87} \\
\bottomrule
\end{tabular}
\end{table}

With pre-computed label embeddings, bi-encoder models maintain near-constant inference speed regardless of label count. GLiNER-bi-edge-v2.0 shows only 5.2\% throughput reduction from 1 to 1024 labels (19.3 to 18.3 examples/s), while gliner\_small-v2.5 experiences 98.7\% degradation (10.7 to 0.14 examples/s). At 1024 labels, the bi-encoder with pre-computation achieves 130× higher throughput than the comparable uni-encoder model. Without pre-computation, bi-encoder models still outperform eBERTa-based uni-encoder variants by 23× at maximum label count, demonstrating inherent architectural efficiency.

The scalability advantage extends across all model sizes. GLiNER-bi-large-v2.0 maintains 2.64 examples/s at 1024 labels with pre-computation, compared to 0.03 examples/s for gliner\_large-v2.5, representing an 88× improvement. For production scenarios with 100 entity types, bi-encoder models achieve 5.3× higher throughput, translating to 1.96M versus 368K predictions per day on a single H100 GPU.

\section{Discussion}\label{sec:discussion}

\subsection{Architectural Trade-offs and Design Rationale}

The bi-encoder architecture addresses fundamental scalability limitations of joint encoding approaches through architectural decomposition. But the central question in bi-encoder design is whether decoupling text and label encoding sacrifices the rich cross-modal interactions that joint encoding provides. Our results suggest this concern is largely unfounded for NER and entity linking tasks: GLiNER-bi-large-v2.0 achieves 61.5\% on CrossNER compared to 60.9\% for gliner\_large-v2.5, indicating that late fusion through dot-product scoring suffices for capturing entity-label alignment. We attribute this to the nature of NER, where entity type semantics are relatively self-contained and do not require deep contextual reasoning about the input text during encoding.

The optional cross-attention fusion layer provides a middle ground for scenarios requiring tighter integration. When enabled, bidirectional attention allows entity-aware text representations and text-aware entity representations while preserving the benefits of separate encoding. This design choice reflects a broader principle: architectural modularity enables practitioners to tune the accuracy-efficiency trade-off for their specific deployment constraints.

\subsection{The Role of Specialized Encoders}

Performance parity with uni-encoder models, despite architectural separation, can be attributed to specialized encoder selection. Sentence transformers optimized for semantic similarity (BGE, MiniLM) provide superior label representation compared to general-purpose encoders. These models are pre-trained on large-scale semantic similarity tasks, making them particularly well-suited for encoding entity type descriptions that must be matched against text spans.

This finding has implications for bi-encoder design more broadly: the choice of label encoder may be as important as the text encoder. Our results show that pairing larger text encoders (ettin-encoder-400m) with correspondingly capable label encoders (bge-base-en-v1.5) yields consistent improvements, while mismatched encoder capacities can lead to bottlenecks. The 194M parameter base variant achieves 98\% of the large model's performance while operating 2.6$\times$ faster, suggesting that careful encoder selection can achieve favorable accuracy-efficiency trade-offs.

\subsection{Domain-Specific Performance Patterns}

Analysis of domain-specific results reveals instructive patterns about the strengths and limitations of bi-encoder architectures. Biomedical benchmarks exhibit strong scaling effects, with GENIA\_NER improving from 50.1\% to 60.1\% (20\% relative gain) between edge and large variants. This suggests that biomedical entity recognition benefits substantially from increased model capacity, likely due to the complex and specialized vocabulary involved.

In contrast, social media datasets (Broad Tweet Corpus, TweetNER7) show relatively stable performance across model sizes (70.0-72.1\% and 36.9-41.7\% respectively). Informal text characteristics—non-standard spelling, abbreviations, and context-dependent meanings—may benefit less from increased model capacity and more from domain-specific pre-training or architectural modifications.

The bi-encoder architecture demonstrates particular advantages on datasets where semantic similarity between labels and text is crucial. Superior performance on GENIA\_NER (60.1\% vs 46.9\%), MultiNERD (68.0\% vs 50.9\%), and Ontonotes (33.3\% vs 22.8\%) compared to uni-encoder models suggests that dedicated label encoders capture entity type semantics more effectively when label descriptions are informative and well-defined.

Conversely, uni-encoder models maintain advantages on established benchmarks like CoNLL 2003 (65.4\% vs 66.5\%), possibly because joint encoding better captures the specific annotation conventions and entity boundary patterns learned during pre-training on similar data distributions.

\subsection{Scalability and Practical Deployment}

The consistent inference speed with pre-computed embeddings transforms large-scale NER deployment from computationally prohibitive to practical. GLiNER-bi-edge-v2.0 shows only 5.2\% throughput reduction from 1 to 1024 labels with pre-computed embeddings, while comparable uni-encoder models experience 98.7\% degradation. This capability proves particularly valuable for several real-world scenarios.

For biomedical applications that rely on extensive ontologies such as UMLS (4M+ concepts), pre-computing and caching entity embeddings enables practical deployment. Traditional approaches would require encoding millions of labels jointly with each input, rendering real-time inference infeasible. With the bi-encoder architecture, label embeddings can be computed once and stored in efficient vector databases, reducing the problem to text encoding followed by nearest-neighbor search.

Enterprise systems requiring dynamic taxonomy updates also benefit substantially. When new entity types are added or existing ones modified, only the affected label embeddings need recomputation, rather than reprocessing the entire model or invalidating cached computations. This modularity aligns well with production ML systems that must evolve continuously.

For production scenarios with 100 entity types, bi-encoder models achieve 5.3$\times$ higher throughput, translating to 1.96M versus 368K predictions per day on a single H100 GPU. This difference can determine whether a system meets latency requirements for user-facing applications or requires costly horizontal scaling.

\subsection{Extension to Entity Linking}

The bi-encoder architecture's natural extension to entity linking represents a significant practical contribution. Entity linking extends NER by disambiguating mentions to specific knowledge base entities—a task that requires candidate retrieval from potentially millions of entities, followed by disambiguation. The bi-encoder design maps directly to this workflow: pre-computed entity embeddings enable efficient candidate retrieval via approximate nearest neighbor search, while the same scoring mechanism used for NER provides disambiguation.

GLiNKER demonstrates this capability through a modular DAG-based pipeline that integrates GLiNER-bi-Encoder as a neural disambiguation component. The framework supports on-the-fly embedding, caching, and database integration for million-scale entity ontologies, addressing a key deployment challenge for knowledge-intensive applications. This unification of NER and entity linking within a single architectural framework simplifies system design and enables end-to-end optimization.

\subsection{Limitations and Future Directions}

Several limitations warrant discussion. First, performance on highly contextual datasets remains challenging. HarveyNER (10.6-15.0\%) and FabNER (22.4-24.3\%) show limited improvements with scaling, indicating that certain entity types require deeper contextual reasoning than current architectures provide. These datasets feature entities whose boundaries and types depend heavily on surrounding context—a scenario where joint encoding may provide fundamental advantages.

Second, the fixed maximum span width constraint ($K=12$ in our experiments) limits recognition of longer entity mentions. It can be sufficient for most NER benchmarks, but specialized domains such as legal or scientific text may contain entity mentions exceeding this limit. While we developed a token-level variant of the bi-encoder architecture, further research is needed to explore its benefits.

Third, while cross-attention fusion is available, our experiments primarily evaluate the base bi-encoder configuration. A systematic study of when and to what extent cross-modal interaction benefits different domains would provide valuable guidance for practitioners.

Future work could explore several promising directions. Contrastive pre-training objectives specifically designed for the bi-encoder NER setting may improve label-text alignment. Integration with retrieval-augmented approaches could enable even larger label vocabularies by dynamically retrieving relevant entity types rather than scoring against all candidates. Finally, extension to multilingual and cross-lingual settings, leveraging multilingual sentence transformers as label encoders, could enable zero-shot NER across languages without parallel training data.

\subsection{Broader Implications}

The success of bi-encoder architectures for NER contributes to a broader trend in NLP toward modular, decomposable systems. Rather than monolithic models that jointly process all inputs, decomposed architectures enable independent component optimization, efficient caching and retrieval, and flexible deployment configurations. This paradigm shift has implications beyond NER—similar architectural principles may benefit other structured prediction tasks where input components have separable semantics.

The 130$\times$ speedup at scale also highlights the importance of computational efficiency in democratizing NLP capabilities. Models that achieve comparable accuracy with dramatically reduced inference costs enable deployment in resource-constrained environments and reduce the environmental impact of large-scale text processing. As NLP systems are deployed more broadly, such efficiency gains become increasingly important.

\section{Conclusion}

We presented GLiNER-bi-Encoder, a bi-encoder architecture for named entity recognition that achieves state-of-the-art zero-shot performance (61.5\% F1 on CrossNER) while demonstrating superior scalability. The architectural separation of text and label encoding enables 130× faster inference than traditional approaches at 1024 entity types, with near-constant complexity when using pre-computed label embeddings.

Experimental evaluation across 26 NER benchmarks confirms that bi-encoder models match or exceed uni-encoder performance while offering transformative computational advantages. The ability to handle thousands of entity types with sub-second latency addresses a critical limitation in current NER systems, enabling deployment in large-scale production environments previously constrained by computational requirements. Moreover, it enables new use cases, such as entity linking.

The models and implementation are publicly available through the GLiNER library and the Hugging Face Model Hub. This work establishes bi-encoder architectures as a practical solution for scalable information extraction, particularly in domains that require extensive entity taxonomies or real-time processing.

\section{Availability}
Models are available for use with the Python library GLiNER at 
 \url{https://github.com/urchade/GLiNER}, and also through the framework UTCA for building pipelines at \url{https://github.com/Knowledgator/utca}. Models can be downloaded from the Hugging Face repository at 
 \url{https://huggingface.co/collections/knowledgator/gliner-bi-v2-68ac5880c610ed907cd68a5a}.

Pre-trained models can be downloaded from the Hugging Face repository at:
\url{https://huggingface.co/collections/knowledgator/gliclass-v3-687a2d211b89659da1e3f34a}

\section{Acknowledgments}
\textit{We sincerely thank Urchade Zaratiana, the creator of the original GLiNER-uni-Encoder, and Tom Aarsen, the maintainer of Sentence Transformers.}

\bibliographystyle{plainnat}  
\bibliography{main}  
\end{document}